\documentclass{article}

\usepackage[final]{corl_2022} % Uncomment for the camera-ready ``final'' version.
\usepackage{graphics} % for pdf, bitmapped graphics files
\usepackage{epsfig} % for postscript graphics files
\usepackage{times} % assumes new font selection scheme installed
\usepackage{amsmath} % assumes amsmath package installed
\usepackage{amssymb}  % assumes amsmath package installed
\usepackage[ruled,vlined,linesnumbered]{algorithm2e}
\usepackage{float}
\usepackage[noend]{algorithmic}
\usepackage{booktabs}
\usepackage{graphicx}
\usepackage[table,xcdraw,dvipsnames]{xcolor}
\usepackage{multirow}
\usepackage{xcolor,pifont}
\usepackage{flushend}
\usepackage{siunitx}
\usepackage[numbers]{natbib}
\usepackage{xcolor}
\usepackage{breqn}
\usepackage{calc}
\usepackage{wrapfig}

\newcommand*{\img}[1]{%
    \raisebox{-.23\baselineskip}{%
        \includegraphics[
        height=\baselineskip,
        width=\baselineskip,
        keepaspectratio,
        ]{#1}%
    }%
}

\let\oldnl\nl% Store \nl in \oldnl
\newcommand{\nonl}{\renewcommand{\nl}{\let\nl\oldnl}}% Remove line number for one line

\title{Residual Skill Policies:\\ Learning an Adaptable Skill-based Action Space for Reinforcement Learning for Robotics}

\author{
    Krishan Rana$^1$, Ming Xu$^1$, Brendan Tidd$^2$, Michael Milford$^1$, Niko S\"underhauf$^1$ \\
    $^1$QUT Centre for Robotics, Queensland University of Technology \\
    $^2$Data61 Robotics and Autonomous Systems Group, CSIRO \\
    \texttt{ranak@qut.edu.au} 
}

\begin{document}
\maketitle

%===============================================================================

\begin{abstract}

Skill-based reinforcement learning (RL) has emerged as a promising strategy to leverage prior knowledge for accelerated robot learning. Skills are typically extracted from expert demonstrations and are embedded into a latent space from which they can be sampled as actions by a high-level RL agent. However, this \textit{skill space} is expansive, and not all skills are relevant for a given robot state, making exploration difficult. Furthermore, the downstream RL agent is limited to learning structurally similar tasks to those used to construct the skill space. We firstly propose accelerating exploration in the skill space using state-conditioned generative models to directly bias the high-level agent towards only \textit{sampling} skills relevant to a given state based on prior experience. Next, we propose a low-level residual policy for fine-grained \textit{skill adaptation} enabling downstream RL agents to adapt to unseen task variations. Finally, we validate our approach across four challenging manipulation tasks that differ from those used to build the skill space, demonstrating our ability to learn across task variations while significantly accelerating exploration, outperforming prior works. Code and videos are available on our project website: \url{https://krishanrana.github.io/reskill}.

% Skill-based reinforcement learning (RL) has emerged as a promising strategy to leverage prior knowledge and temporal abstraction in RL for accelerated robot learning. However, for the skills to be effective, they need to be extracted from diverse expert datasets that capture optimal behaviours for the downstream tasks. Such datasets are not always readily available and can limit the generalisation of skill-based RL to task or environmental variations. This work proposes a novel skill adaptation framework that can effectively utilise a limited number of skills for solving a wide range of unseen tasks. We re-purpose existing handcrafted controllers as readily available sources of skills for robotics. These skills may not necessarily cover the entire range of behaviours required for the downstream tasks and may be suboptimal. To make effective use of these skills, we propose a low-level residual policy for fine-grained \textit{skill adaptation} to task variations. Finally, to accelerate downstream RL, we propose a state-conditioned skill-space that biases the agent to explore only the \textit{meaningful} skills based on the current state. We empirically show that this formulation allows RL agents to effectively leverage limited and sub-optimal skills datasets to accelerate learning without crippling their performance on unseen tasks.
\end{abstract}


%===============================================================================

\section{Introduction}

Humans are remarkably efficient at learning new behaviours. We can attribute this to their ability to constantly draw on relevant prior experiences and adapt this knowledge to facilitate learning. Building on this observation, recent works have proposed various methods to incorporate the use of prior experience in deep reinforcement learning (RL) to accelerate robot learning.

One approach that has gained attention recently is the use of \textit{skills} which are short sequences of single-step actions representing useful and task-agnostic behaviours extracted from datasets of expert demonstrations \cite{merel2018neural, Lynch2019LearningLP, pertsch2020spirl}. In the manipulation domain, for example, such skills could include \textit{move-left}, \textit{grasp-object} and \textit{lift-object}. These skills are typically embedded into a latent space, forming the action space for a high-level RL policy. While yielding significant improvements over learning from \textit{scratch}, there are still outstanding challenges for training RL agents using this latent \textit{skill space}. 

\begin{figure}[t]
\centering
\includegraphics[width=0.95\textwidth]{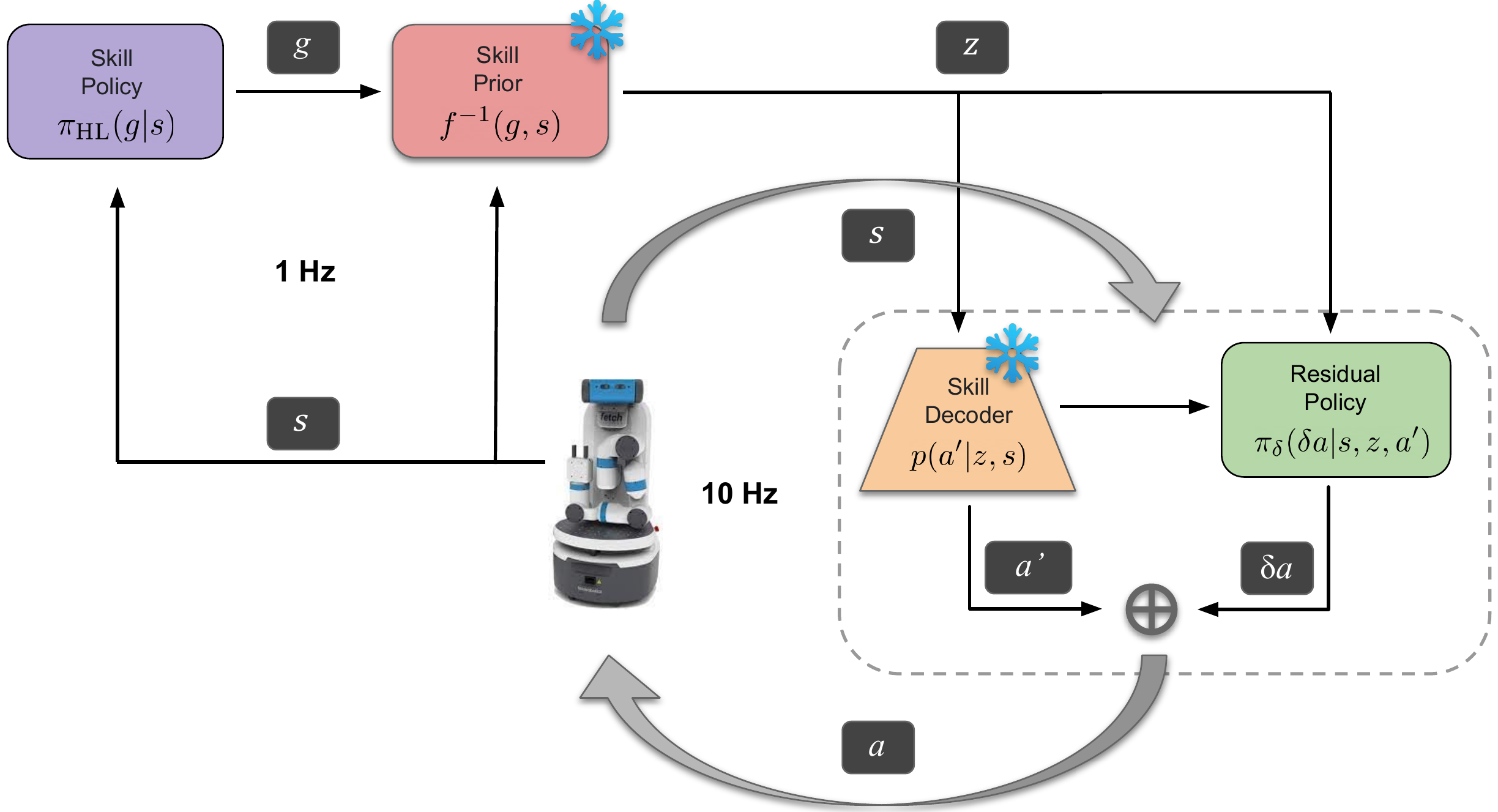}
\caption{\textbf{Residual Skill Policies (ReSkill).} A skill-based learning framework for robotics. The skill prior transforms the agent's action space to a state-conditioned skill space using normalising flows, where only the relevant skills for the current state are explored. The residual policy allows for fine-grained adaptation of the skills to environment variations and unseen tasks. The \img{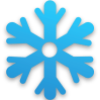} symbol signifies that the pre-trained weights for these skill modules are frozen during downstream task learning.
}
\label{rsp}
\end{figure}

Firstly, na\"ive exploration by sampling randomly over all skills can be extremely inefficient~\cite{Lynch2019LearningLP, pertsch2020spirl, singh2021parrot}. Typical skill spaces encode an expansive set of generic skills, with only a small fraction of skills relevant for execution in a given robot state. Furthermore, the set of skills relevant to a given state do not typically cluster in the same neighbourhood in the skill space, i.e.~the conditional density is \textit{multi-modal}, making targeted sampling challenging. We address this issue of inefficient exploration by learning a \textit{state-conditioned skill prior} over the skill space, which directly biases the agent's exploration towards sampling only those skills that are deemed relevant in its current state. Unlike previous works \cite{pertsch2020spirl}, our skill-prior captures the complex multi-modality of the skill space allowing us to sample directly from this density during exploration.

Secondly, current approaches to skill-based RL assume that the skills are optimal and that downstream tasks are drawn from the same distribution used to create the skill space. Consequently, skill-based learning methods have limited generality and adaptability to task variations. For example, if the available skills were extracted solely from demonstrations of block manipulation on an empty table, learning a downstream task with obstacles, object variations, or different friction coefficients may not be possible. In this case, using a non-exhaustive skill space as an action space can cripple an RL agent's learning ability. We address this by introducing a low-level residual policy, which enables fine-grained \textit{skill adaptation} of available skills to variations in the task not solvable by simply composing the available skills. This relaxes the need for exhaustive and expert demonstration datasets to one that can be extracted from existing classical controllers, which are a cheap and readily available source of demonstration data for robotics.

We summarise the main contributions of our approach denoted as Residual Skill Policies (ReSkill) as follows: \textbf{(1)} we propose a novel state-conditioned skill prior that enables direct sampling of relevant skills for guided exploration, \textbf{(2)} we introduce a low-level residual policy that can adapt the skills to variations in the downstream task, and \textbf{(3)} we demonstrate how our two contributions both accelerate skill-based learning and enable the agent to attain higher final performance in novel environments across four manipulation tasks, outperforming all baseline methods.

\section{Related Work}

% \subsection{Skill-Based RL}
% karols paper - learn it with rl
% karls paper - learn skills from offline datasets
% parrot paper
% play paper
% npmp paper for cloning plicies

\paragraph{Skill-Based RL} This has emerged as a promising approach to effectively leverage prior knowledge within the RL framework by embedding reusable skills into a continuous skill space via stochastic latent variable models \cite{merel2018neural, Lynch2019LearningLP, pertsch2020spirl, hausman2018learning, merelcatch}. Skill-based learning provides temporal abstraction that allows for effective exploration of the search space, with the ability to draw on previously used behaviours to accelerate learning. However as described previously, the skill space can be difficult to explore. \citet{pertsch2020spirl} proposed the use of a prior over skills to help better guide exploration. They learn a unimodal Gaussian prior that approximates the density of relevant skills for a given state and regularise the downstream RL agent towards the prior during training to indirectly bias exploration. \citet{singh2021parrot} proposed a more expressive strategy for learning these priors via the use of state-conditioned generative models. While this was conducted at the single-step action level, they show that they can directly bias exploration by sampling relevant actions from this density. In this work, we extend this idea to skill-based action spaces. 

% While overall an attractive strategy to bringing general purpose and reusable prior experience to reinforcement learning, skill based learning is heavily dependent on obtaining a diverse repertoire of general-purpose and expert skills. The lack thereof can greatly impact the final performance of the downstream learning algorithm, as well as the generality of the skills modules to a broad range of unseen tasks. In this work, we allow the RL agent to directly adapt these skills to downstream task variations. This reduces the need for both exhaustive and expert skills datasets which are difficult to obtain.

% \todo{make a comment about the structure of a latent space; difficult to interpret/sample from; regions may not have encoded skills in some regions, random exploration is hard; skill space is designed to be very general with skills across a wide range of tasks resulting in a wide range of skills some of which are irrelevant for particular tasks; Guassian prior does that capture the complex distribution followed by the actually skill embeddings.}

% \subsection{Combining RL with Classical Control}
\paragraph{Combining RL with Classical Control} Our work is closely related to methods that seek to combine the strengths of classical control and RL for accelerated learning. In contrast to expert datasets or human demonstrators, classical control systems are cheap and readily available for most robotics tasks. \citet{johannink2018residual} proposed the residual reinforcement learning framework in which an RL agent learns an additive residual policy that modifies the behaviours of a predefined handcrafted controller to simplify exploration. Other works in this area have shown that combining the two modes of control allows for sample efficient initialisation \cite{rana_mcf, uchendu2022jump, xie2018learning}, temporal abstraction for accelerated learning \cite{relmogen}, and safe exploration \cite{rana2021bayesian}. While strongly motivating the re-use of existing handcrafted controllers, these methods are restricted to solving only one particular task for which the controller was designed, requiring a new controller for each new task. This limits the generality of these approaches to novel tasks. In this work, we decompose these controllers into task-agnostic skills and explore how they can be repurposed for learning a wide range of tasks.

% \subsection{Hierarchical RL}

\paragraph{Hierarchical RL (HRL)} These methods enable autonomous decomposition of challenging long-horizon decision-making tasks for efficient learning. Typical HRL approaches utilise a high-level policy that selects from a set of low-level temporally extended actions or options \cite{sutton1999between} that provide broader coverage of the exploration space when executed \cite{steccanella2020hierarchica}. Prior work in hierarchical reinforcement learning can be divided into two main categories: learning low-level and high-level policies through active interaction with an environment \cite{Bacon_Harb_Precup_2017, Kupcsik2013DataEfficientGO, Heess2016LearningAT, Haarnoja2018LatentSP, hac}, and learning options from demonstrations before re-composing them to perform long-horizon tasks through RL \cite{shankar2019discovering, Krishnan2017DDCODO, sharma2018directedinfo}. Our work shares similarities with both strategies. We follow the latter approach by pre-training a low-level skill module from demonstration data on which a high-level policy operates, while additionally learning a low-level residual policy online for fine-grained skill adaptation.

% Most similar to the ideas presented in skill-based RL is the work by Sharma \textit{et al.} \cite{sharma2018directedinfo} who naturally decompose demonstrations into sub-task policies which can later be recomposed by a high level policy.  

% \subsection{Meta RL}

% Our goal in this work is to re-purpose existing handcrafted controllers as skills in order to accelerate RL when learning a wide range of novel tasks. Meta-learning \cite{duan2016rl, wang2016learning, rakelly2019efficient} approaches similarly seek to extract useful information from previous experience to improve the learning efficiency for unseen tasks. While meta-learning provides an appealing and principled framework to accelerate acquisition of future tasks, they are typically restricted to short-horizon, dense-reward tasks \cite{finn2017model} and require memory models in order for the policy to internalise the dynamics between states, rewards, and actions in the current MDP and adjust its strategy accordingly. Our approach is orthogonal and more relaxed, via the use of skills and state-conditioned skill-spaces, which capture the prior knowledge required to accelerate the learning of new tasks with the ability to solve long-horizon tasks with sparse rewards.

\section{Problem Formulation}

We focus on skill-based RL which leverages \textit{skills} as the actions for a high-level policy $\pi_{\text{HL}}$, where a skill $\textit{\textbf{a}}$ is defined as a sequence of atomic actions $\{a_t,...,a_{t+H-1}\}$ over a fixed horizon $H$. Skills are typically embedded into a continuous latent space $\mathcal{Z}$ from which $\pi_{\text{HL}}$ can sample during exploration \cite{merel2018neural, Lynch2019LearningLP, pertsch2020spirl}. Latent \textit{skill spaces} expressively capture vast amounts of task-agnostic prior experience for re-use in RL. However, they pose several challenges that can negatively impact downstream learning. Firstly, current methods rely on expert and exhaustive demonstration datasets that capture skills for a wide range of potential task variations. Such datasets can be arduous and expensive to obtain. Secondly, as the skill space abstracts away the RL agent's atomic action space, the generality of skill-based RL is greatly limited to tasks that closely match the distribution of those captured by the dataset. This poses a trade-off between generality and sample efficiency in downstream learning \cite{gehring2021hierarchical}. Furthermore, latent skill spaces are expansive and can be arbitrarily structured, making random exploration difficult. Learning structure within this space can better inform exploration to accelerate downstream learning.

The downstream task is formulated as a Markov Decision Process (MDP) defined by a tuple $(\mathcal{S,A,T,R},\gamma)$ consisting of states, actions, transition probabilities, rewards and a discount factor. We learn the high-level policy $\pi_{\text{HL}}(z|s)$ to select the appropriate latent skills $z$ in a given state $s$ to maximise the discounted sum of $H$-step rewards $J=\mathbb{E}_{\pi_{\text{HL}}}[\sum_{t=0}^{T-1} \gamma^{t} \tilde{r}]$, where $T$ is the episode horizon, $\tilde{r}$ is the $H$-step reward defined as $\tilde{r}=\sum_{t=1}^{H}r_t$ and $r_t$ is the single-step reward.

\begin{figure}[t!]
  \centering
  \includegraphics[width=\textwidth]{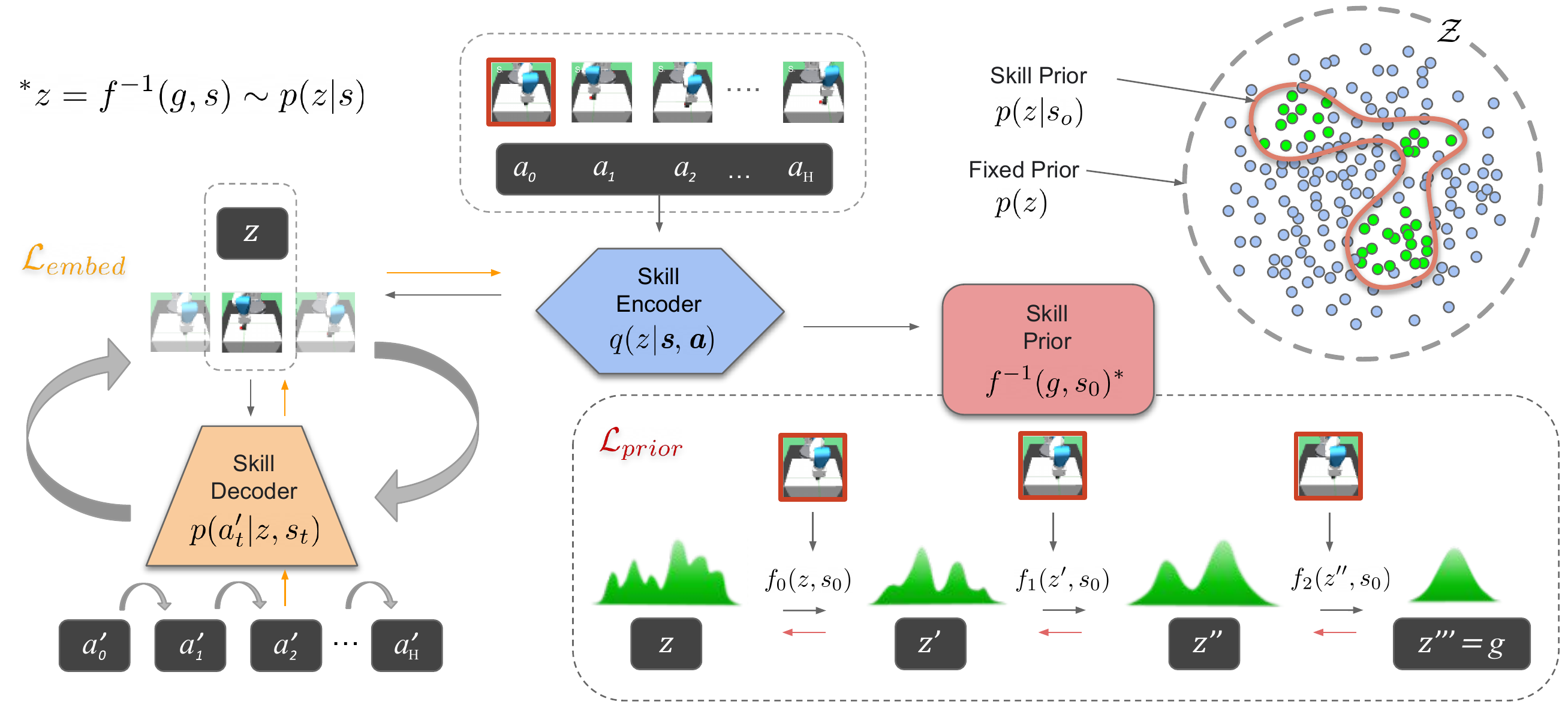}
  \caption{\textbf{Schematic for learning the state-conditioned skill space.} We train a VAE \textit{embedding module} which encodes skills into a latent embedding space $\mathcal{Z}$. This module is comprised of an encoder and a closed-loop skills decoder, where the decoder recovers atomic actions from a latent skill embedding $z$. The \textit{skill prior} module learns the state-conditioned density of useful skills based on the dataset. This conditional density is multi-modal in the skill space and we estimate it using normalising flows. Both modules are trained jointly, with coloured arrows illustrating the gradient flow between them. }
    \label{method}
\end{figure}

\section{Approach}

We aim to re-purpose existing classical controllers as skills to facilitate RL in solving new tasks. We decompose demonstration trajectories produced by these controllers into task-agnostic skills and embed them into a continuous space $\mathcal{Z}$ using stochastic latent variable models. To facilitate exploration for downstream RL in this skill space, we propose a state-conditioned \textit{skill prior} $p(z|s_0)$  over skills that directly biases the agent towards only sampling skills relevant to a given state based on prior experience. Finally, to enable truly general-purpose learning with the skill space, we propose a low-level residual policy $\pi_{\delta}(\delta a|\cdot)$ that can adapt the embedded skills to task variations not captured by the demonstration data. This gives the RL agent access to its single-step action space and relaxes the need for both expert and exhaustive datasets.

Our approach can be decomposed into three sub-problems: (1) skill extraction from existing controllers, (2) learning a skill embedding and skill prior, and (3) training a hierarchical RL policy within this skill space with a low-level, residual skill adaptation policy. We discuss each of these steps in more detail below.

\subsection{Data Collection}\label{ssec:data}

We re-purpose existing handcrafted controllers as sources of demonstration data for basic manipulation tasks involving \textit{pushing} and \textit{grasping} an object on an empty table. While these controllers are relatively simple, the trajectories produced possess a vast range of skills that can be recomposed to solve more complex tasks. We note here that these trajectories do not need to be optimal given our adaptable downstream RL formulation which we will describe in more detail in the following sections. The trajectories consist of state-action pairs and we perform unsupervised skill extraction by randomly slicing a $H$ length segment from them. We utilise both the extracted sequence of actions $\textit{\textbf{a}}$ as well as the corresponding states ${\textit{\textbf{s}}}$ to learn the state-conditioned skill space described in the next section. The state vector comprises of joint angles, joint velocities, gripper position and object positions in the scene, while the actions are continuous 4D vectors representing end-effector and gripper velocities in Cartesian space. More details about data collection and the specific handcrafted controllers used can be found in Appendix \ref{app:data}.

\subsection{Learning a State-Conditioned Skill Space for RL}

% TODO: High level skills policy has two components: s to n and n to z. n to z is trained offline using RNVP, s to n is trained after using MDP stuff. High level skills policy is one module with two components? Or we have a high level skills policy and a skills prior? About demarcation and semantics and naming.

Learning a state-conditioned skill space consists of two key steps: (1) embedding the extracted skills into a latent space; (2) learning a state-conditioned prior over the skills which we can sample from during exploration. Figure \ref{method} provides a summary of our approach and we describe each component in detail below.

\subsubsection{Embedding the skills}

We follow the approach proposed in prior works \cite{ merel2018neural, pearce2018uncertainty}, where skills $\textit{\textbf{a}}$ are embedded into a latent space using a variational autoencoder (VAE) \cite{Kingma2014}. The VAE comprises of a probabilistic encoder $q_\phi$ and decoder $p_\theta$ with network weights $\phi$ and $\theta$ respectively, which maps a skill action sequence to a latent embedding space and vice versa. Our encoder $q_\phi(z \mid \textit{\textbf{s}}, \textit{\textbf{a}})$ jointly processes the full state-action sequence, while our decoder $p_\theta(a_t \mid z, s_t)$ reconstructs individual atomic actions conditioned on current state $s_t$ and skill embedding $z$. The loss function for our VAE for a single training example is given by
\begin{equation}\label{eq:vaerecon}
\begin{small}
\mathcal{L}_{embed} = \mathbb{E}_{q_\phi(z \mid \text{\textit{\textbf{s}}}, \text{\textit{\textbf{a}}})}\left[ \sum_{t=0}^{H-1} \log p_\theta\left(a_{t} \mid z, s_{t}\right)\right]-\beta D_{KL}(q_\phi\left(z \mid \textit{\textbf{s}}, \textit{\textbf{a}} \right) \parallel p(z)),
\end{small}
\end{equation}

where $p(z)\sim \mathcal{N}(0, I)$ and $\beta\in[0, 1]$ is the weighting for the regularisation term \cite{Kingma2014}.  The \textit{closed-loop} nature of the decoder considers the current robot state $s_t$ as well as sampled skill $z$ when recovering actions, which was found to significantly improve performance in downstream learning \cite{pertsch2021skild}, particularly in dynamic environments.

\begin{wrapfigure}[20]{r}{0.45\textwidth} %17
 \vspace{-0.95cm}
  \centering
  \includegraphics[width=0.45\textwidth]{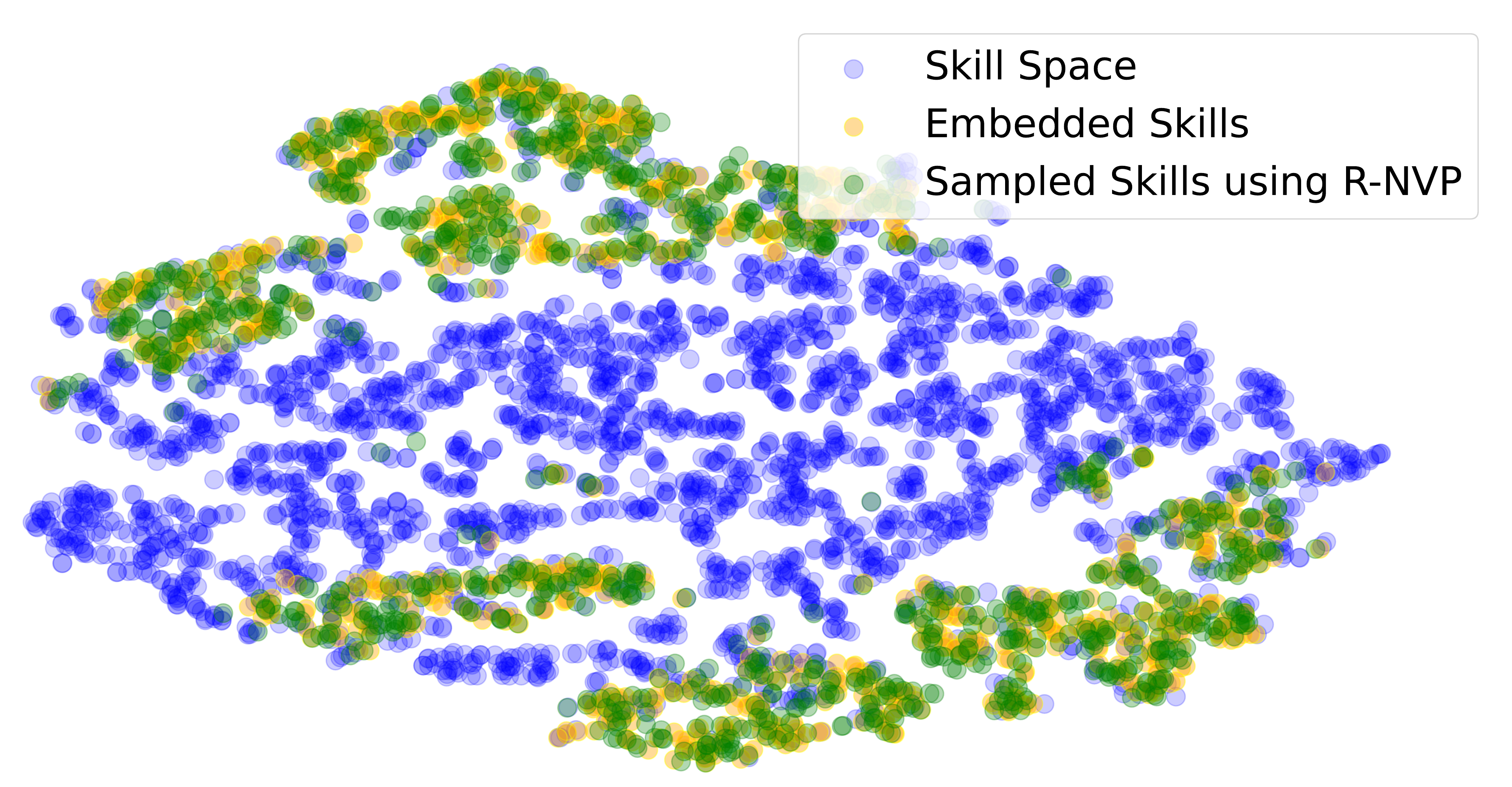}
%   \vspace{-0.5cm}
  \caption{\textbf{t-SNE of the skill embedding space.} Blue denotes samples drawn from the VAE prior $p(z)$. Yellow denotes embedded skills drawn from the dataset. Green denotes samples using our trained skill prior. Note the multi-modality across the latent space. The skill prior samples from regions containing embedded and relevant skills, reducing the chances of sampling irrelevant regions (blue) not containing any meaningful skills.}
  \label{fig:z_space}
\end{wrapfigure}

\subsubsection{Learning the state-conditioned skill prior}\label{sssec:skillprior}
The nature of the VAE reconstruction loss described in Equation \eqref{eq:vaerecon} allows us to sample skills from the latent space by using the prior $p(z)\sim \mathcal{N}(0, I)$. However, sampling directly from the prior can be inefficient, since the samples will range across the full, expansive set of available skills as well as regions that may not constitute any embeddings as shown in Figure \ref{fig:z_space}. In reality, for a particular robot state, only a small subset of these skills are relevant for exploration. To address this inefficiency, we propose to learn a \textit{state-conditioned skill prior} over the latent embedding space, which allows us to more frequently sample the \textit{relevant skills} likely to be useful in a given state, thus accelerating exploration.

Mathematically, we are looking to learn a conditional probability density over the latent skill space $p(z|s_0)$ that we can draw samples from when training $\pi_{HL}$. This conditional density is typically highly multi-modal, i.e.~for a given robot state, many skills can be relevant that may be far apart in the topology of the latent skill space. Prior works \cite{pertsch2020spirl} propose a simple unimodal density (e.g. Gaussian) which does not sufficiently capture this multi-modality and hence is unsuitable to directly sample skills from.

One of our contributions is around gracefully handling this multi-modality by using the real-valued non-volume preserving transformations (real NVP) method by \citet{dinh2016density} based on normalising flows \cite{tabak2013family}. Conditional real NVP \cite{ardizzone2018analyzing} learns a mapping $f:\mathcal{Z} \times \mathcal{S} \rightarrow \mathcal{G}$ which we can use to generate samples from $p(z|s)$. Spaces $\mathcal{Z}$ and $\mathcal{G}$ are identical, i.e.~$\mathcal{Z}\cong\mathcal{G}\cong\mathbb{R}^d$ and furthermore $f$ is bijective between $\mathcal{Z}$ and $\mathcal{G}$ for a fixed state $s$. Sampling is achieved as follows: first sample $g \sim p_{\mathcal{G}}(g)$ from a simple distribution (e.g. $p_{\mathcal{G}}(g)\sim\mathcal{N}(0, I)$) and then transform $g$ into skill space $\mathcal{Z}$ using $f^{-1}$, i.e.~$f^{-1}(g, s)\sim p(z|s)$. Since $f$ fully defines $p(z|s)$, we henceforth refer to $f$ as the skill prior.

We refer the reader to \citet{dinh2016density} for more information about real NVP. We train our state-conditioned skill prior jointly with our VAE embedding module. Figure \ref{fig:z_space} illustrates the multi-modality present in the latent skill space and how our skill prior captures this. Implementation details are provided in Appendix \ref{app:generative}.

\subsection{Reinforcement Learning in a State-Conditioned Skill Space}

Once trained, the decoder and skill prior weights are frozen and incorporated within the RL framework. Our high-level RL policy denoted $\pi_{\text{HL}}(g|s)$ is a neural network that maps a state to a vector $g \in G$ in the domain of the skill prior transformation $f^{-1}$ defined in Section \ref{sssec:skillprior}. The skill prior $f^{-1}$ is then used to transform $g$ to latent skill embedding $z$. Our use of normalising flows for our skills prior is motivated by \cite{singh2021parrot}, who hypothesised that the bijective nature of $f$ allows the RL agent to retain full control over the broader skill space $\mathcal{Z}$: for any given $z$, there exists a $g$ that generates $z=f^{-1}(g,s)$ in the original MDP. This means that all skills from $\mathcal{Z}$ can be sampled by the skill prior with non-zero probability.

Once a latent skill $z$ is attained by the high-level agent, the closed-loop decoder can reconstruct each action $a'$ sequentially conditioned on the current state for the skill horizon $H$. To increase the diversity of downstream tasks that can be learned using the fixed skill space, we introduce a low-level residual policy $\pi_\delta$ that can adapt the decoded skills at the executable action level, providing the hierarchical agent with full control over the MDP. We condition this residual policy on the current state $s$, selected skill $z$, and the action proposal generated by the skill decoder $a'$. The produced action $\delta a \sim \pi_\delta(\delta a | s, z, a')$ is added to the decoded action $a'$ before being executed on the robot. This low-level system is operated at a frequency of $H$ Hz based on the skill horizon, while the outer loop operates at 1 Hz. Figure \ref{rsp} summarises our Residual Skill Policy (ReSkill) architecture and pseudo-code for action selection is provided in Algorithm \ref{algo:act_selection}. Given the non-stationarity imposed by the low-level residual policy being trained jointly with the high-level policy, we utilise an on-policy RL algorithm for training the hierarchical agent. Implementation details are provided in Appendix \ref{app:rl}.

\begin{algorithm}[t]
\caption{Action Selection}
\nonl\textbf{Given:} high-level policy $\pi_{\text{HL}}$, residual policy $\pi_{\delta}$, skill decoder $p$, skill prior $f$\\
\nonl\textbf{Inputs:} state $\textit{s}_t$
\begin{algorithmic}[1]
  \STATE $g \sim \pi_{\text{HL}}(g|s_t)$\COMMENT{\textcolor{gray}{sample a vector $g$ from the high-level policy}}
  \STATE $z \leftarrow f^{-1}(g,s_t)$ \COMMENT{\textcolor{gray}{map $g$ to a state-relevant skill using the skill prior}}
  \FOR{skill horizon $H$}
  \STATE $a' \leftarrow p(a'|s_t, z)$ \COMMENT{\textcolor{gray}{decode the skill to a single-step action using the skill decoder}}
  \STATE $\delta a \sim \pi_{\delta}(\delta a|s_t, z, a')$ \COMMENT{\textcolor{gray}{sample a corrective action from the residual policy}}
  \STATE $a_t = a' + \delta a$ \COMMENT{\textcolor{gray}{sum the decoded and residual action}}
  \STATE \textbf{return} action $\textit{a}_t$ \COMMENT{\textcolor{gray}{execute the combined action in the environment and continue}}
  \ENDFOR
  \end{algorithmic}
\label{algo:act_selection}
\end{algorithm}

\section{Experiments}

\begin{figure}[t]
  \centering
  \includegraphics[width=\textwidth ]{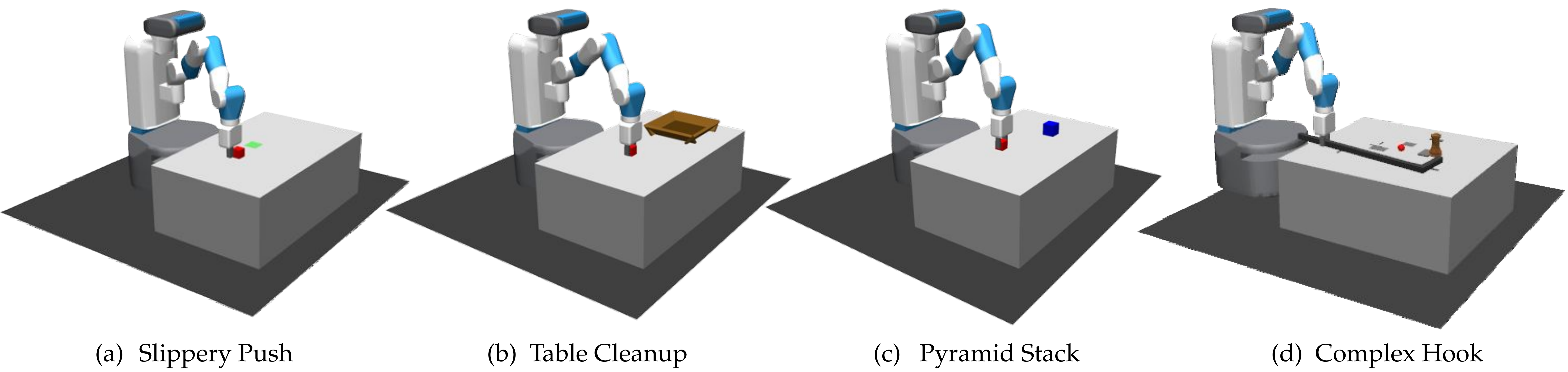}
  \caption{\textbf{Downstream Tasks.} Robotic manipulation tasks involving a 7 DoF manipulator arm from Fetch Robotics. Each task exhibits a physical or dynamical variation to the environment that was not present in the skill extraction environment.}
  \label{robot_envs}
\end{figure}

We evaluate our approach in a robotic manipulation domain simulated in MuJoCO \cite{todorov2012mujoco}, involving a 7-DoF Fetch robotic arm interacting with objects placed in front of it as shown in Figure \ref{robot_envs}. Evaluation is performed over four downstream tasks, each one not solvable purely with the prior controllers used for data collection as described in Appendix \ref{app:data}. Each task is adapted from the residual reinforcement learning set of environments proposed by Silver \textit{et al.} \cite{silver2018residual}. We describe each of the downstream tasks in detail in Appendix \ref{app:tasks}.

% We evaluate our approach in a robotic manipulation domain simulated in MuJoCO \cite{}, involving a 7-DoF Fetch robot \cite{} interacting with blocks placed in front of it as shown in Figure \ref{envs}. The downstream learning task introduces additional complexity not entirely solvable with the simplistic controllers used for data collection, and requires the RL agent to recompose and adapt these behaviours in order to succeed. These tasks are described as follows:\\
% \noindent\textbf{Slippery Push:} The agent is required to push a block to a given goal, however we lowered the friction of the table surface from that seen in the training push task.\\
% \textbf{Table Cleanup:} We introduce a rigid tray object in the scene, into which the agent must place the given block. The tray is not observed by the agent and the edges act as an obstruction to the decoded skills. \\
% \textbf{Pyramid Stack: } Blocking stacking task, where the agent is rewarded for placing the smaller block onto a larger block. The skills dataset does not contain movements of the arm above the height of the larger block.\\

\noindent We compare our approach against several state-of-the-art RL methods; these methods either operate directly on the single-step action space or train agents using a skills-based approach. The specific methods we compared against are as follows:

\begin{enumerate}
    \item \textbf{Scripted Controller:} The average performance of the handcrafted controllers used for data collection across the 4 environments.
    \item \textbf{Behavioural Cloning (BC) + Fine-Tuning:} Trains an agent by first using supervised learning on state-action pairs from the training set, followed by RL fine-tuning.
    \item \textbf{SAC} and \textbf{PPO:} Trains an agent using Soft Actor-Critic \cite{haarnoja2018soft} and Proximal Policy Optimisation \cite{schulman2017proximal} respectively with random weight initialisation.
    \item \textbf{HAC:} Trains an agent using Hierarchical Actor-Critic \cite{hac} using a hierarchy of 2 for all tasks.
    \item \textbf{PaRRot:} Leverages flow-based density estimation to map the agent's action space to single-step actions that are likely to be useful in a given state \cite{singh2021parrot}.
    \item \textbf{SPiRL:} A hierarchical skill-based approach utilising KL regularisation of the RL policy towards a Gaussian skill prior \cite{pertsch2020spirl} 
    \item \textbf{ReSkill (No Skill Prior):} Ablation study where the high-level policy samples skills directly using $\pi_{\text{HL}}({z|s})$ with no skill prior guidance.
    \item \textbf{ReSkill (No Residual):} Ablation study where we remove the low-level residual policy for adapting skills.
\end{enumerate}

% \textbf{Scripted Controller:} The average performance of the handcrafted controllers used for data collection across the 3 environments.\\
% \textbf{Behavioural Cloning (BC) + Fine-Tuning} Learns a policy by first using supervised learning on state-action pairs from the training set, followed by RL fine-tuning.\\
% \textbf{SAC:} Trains an agent using Soft Actor Critic \cite{haarnoja2018soft} with random weight initialization.\\
% \textbf{PPO:} Trains an agent using Proximal Policy Optimisation \cite{schulman_proximal_2017}  with random weight initialization.\\
% \textbf{PaRRot:} Leverages flow-based density estimation to map the agent's action space to atomic actions that are likely to be useful in a given state \cite{singh2021parrot}.\\
% \textbf{SPiRL:} A skill-based approach utilising KL regularisation of the RL policy towards a Gaussian skill prior \cite{pertsch2020spirl} \\
% \textbf{ReSkill (No Skill Prior:)} Ablation study where the high-level policy samples from $\mathcal{Z}$ directly with no skill prior.\\
% \textbf{ReSkill (No Residual: )} Ablation study where we remove the low-level residual policy for adapting skills.\\
% Method 1 is a baseline, methods 2-5 operate directly on atomic actions and method 6 is a state-of-the-art skills-based method.

\section{Results}

\begin{figure}[t]
  \centering
  \includegraphics[width=\textwidth]{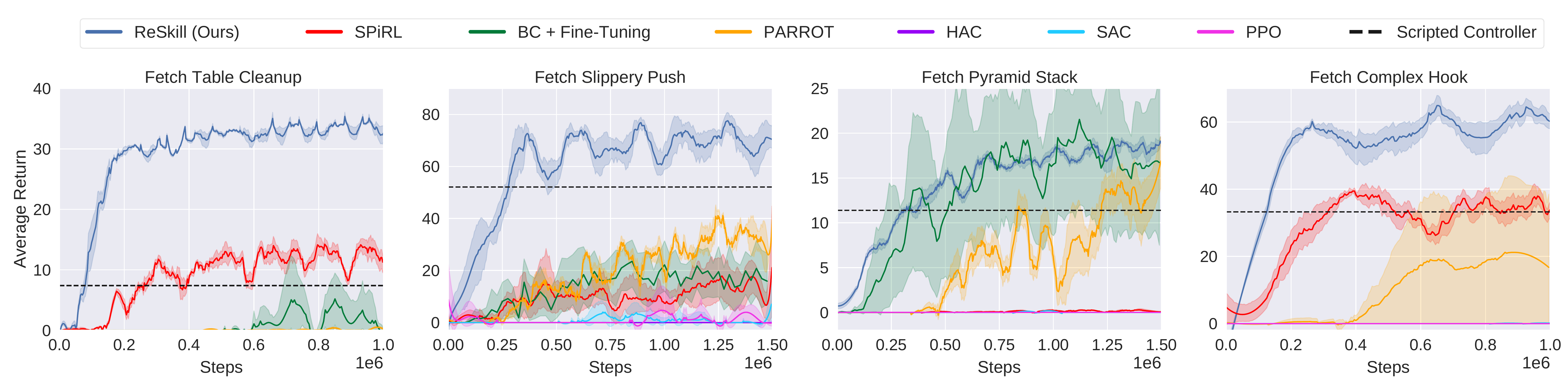}
  \caption{\textbf{Training Performance.} Average training performance across 5 random seeds for the different tasks. ReSkill outperforms all the baselines in both sample efficiency and convergence to the highest-performing final policy.}
\label{curves}
\end{figure}

\begin{figure}[t]
  \centering
  \includegraphics[width=\textwidth]{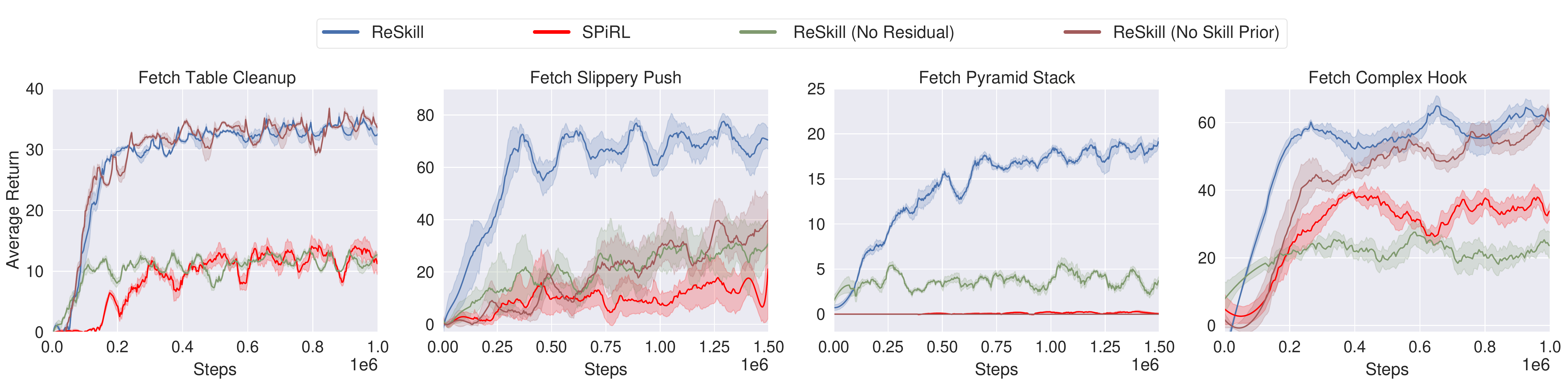}
  \caption{\textbf{Ablation Study.} Analysing the impact of the residual and skill prior on the average training performance across the 4 tasks. The skill prior plays an important role in accelerating learning with the more difficult tasks, while the residual is important for attaining higher rewards by adapting the skills to the variations in the training environment.}
\label{ablation}
\end{figure}

The results across each task are summarised in Figure \ref{curves}. Our approach, ReSkill, outperforms all the baselines, attaining the highest performing policy given its ability to adapt to task variations while additionally converging with superior sample efficiency given the guidance of the state-conditioned skill prior. Note, the baseline scripted controllers are sub-optimal across all four tasks as indicated by the dashed black line. SAC, PPO and HAC fail to learn altogether, given the inability of random exploration alone to yield any meaningful behaviours in sparse reward settings. Results are incrementally better using BC initialisation and fine-tuning; however, the agent still attains sub-optimal performance with high variance across the trials. A critical difference between PaRRot and ReSkill is the use of skills in ReSkill instead of single-step actions. The temporal abstraction provided by skills allows ReSkill to significantly outperform PaRRot in sample efficiency across all tasks. Further, we note that PaRRot struggles to learn a high-performing policy in the \textit{Slippery-Push} and \textit{Complex-Hook} domain which we could attribute to its inability to deal with the variations in the downstream task given the constrained action space. SPiRL accelerates learning, yielding better sample efficiency than all atomic-action-based methods on \textit{Table-Cleanup} and \textit{Complex-Hook}. However, it still fails to attain high final policy performance as it only composes high-level skills without the ability to adapt them to task variations, as shown by ReSkill.

We additionally provide an ablation study in Figure \ref{ablation} to better understand the impact of the skill prior and residual agent on the overall skill-based learning architecture. ReSkill (No Skill Prior) demonstrates the importance of our sampling-based prior in accelerating learning and enabling our agent to learn altogether in the more complex stacking task. The ReSkill (No Residual) plots demonstrate the importance of the residual in enabling the agent to attain high final performance by re-introducing access to the single-step action space for fine-grained control. This allows the agent to appropriately balance the trade-off between generality and sample efficiency when leveraging skills in RL. Furthermore, comparing ReSkill (No Residual) to SPiRL, we note that our approach attains faster convergence to the final policy. We can attribute this to the nature of our skill prior which enables the agent to directly sample and execute state-relevant skills from the early stages of training. In contrast, SPiRL relies on an indirect bias to the agent's exploratory actions via KL-regularisation towards a unimodal skill prior. This requires an initial "burn-in" phase before the policy is appropriately regularised towards the skill prior in order for the agent to start sampling relevant skills. We provide a more detailed ablation study of the impact of our proposed skill-prior in Appendix \ref{skillpablation}.

\subsection{Limitations}

While we demonstrate significantly faster learning and better asymptotic performance than all the baseline methods, ReSkill still requires thousands of environment samples before converging to a high success rate. This limited our ability to demonstrate our method on a physical robot. An interesting future direction to explore would be to integrate our approach with an off-policy RL algorithm such as SAC in order to make better use of exploratory experience for improved sample efficiency. This, however, would require careful consideration of the non-stationarity imposed by the low-level residual component. Secondly, while we note that the skill prior used in this work is bijective and should theoretically allow the agent to access the entire range of skills in the larger skill space, the likelihood of selecting useful skills not seen in the training dataset becomes very low. This would particularly be an issue when there is a significant mismatch between the training dataset and the downstream task. We note here that the residual policy could play a major in addressing this, however, we leave a thorough evaluation of this to future work. Finally, learning the state-conditioned skill space requires two modules, specifically the VAE embedding module and flows-based skill prior module. Future work will explore using a single generative model such as a conditional VAE \cite{NIPS2015_8d55a249} to jointly handle the tasks of both modules.

\section{Conclusion}
This work proposes Residual Skill Policies (ReSkill), a general-purpose skill-based RL framework that enables effective skill re-use for adaptable and efficient learning. We learn a high-level skill agent within a state-conditioned skill space for accelerated skill composition, as well as a low-level residual agent that allows for fine-grained skill adaptation to variations in the task and training environment. Additionally, we show that our framework allows us to re-purpose a handful of scripted controllers to solve a wide range of unseen tasks. We evaluate ReSkill across a range of manipulation tasks and demonstrate its ability to learn faster and achieve substantially higher final returns across all task variations than the baseline methods. We see this as a promising step towards the effective re-use of prior knowledge for efficient learning without limiting the generality of the agent in learning a wide range of tasks.

% Future work can explore how to continually embed new skills within the latent space, particularly those composite skills resulting from combining the residual policy and skill decoder. While the state conditioning on the skill-space did narrow down and accelerated exploration, it limited the generality of the decoder to changes in the state information. Research surrounding information-asymmetry \cite{galashov2019information} or potentially exploring alternative task agnostic conditioning could be explored to learn a truly general-purpose skills module.

%===============================================================================

\clearpage
% The acknowledgments are automatically included only in the final and preprint versions of the paper.
\acknowledgments{ K.R., M.X., M.M. and N.S. acknowledge the continued support from the Queensland University of Technology (QUT) through the Centre for Robotics. This research was additionally supported by the QUT Postgraduate Research Award. The authors would like to thank Robert Lee, Jad Abou-Chakra, Fangyi Zhang, Vibhavari Dasagi, Tom Coppin and Jordan Erskine for their valuable discussions towards this contribution. We would also like to thank the reviewers for their insightful and constructive comments towards the significant improvement of this paper.}

%===============================================================================

\bibliography{references.bib,manual.bib,manual_bib2.bib}  % .bib

\newpage
\appendix

\section*{Appendix}

\section{Implementation Details}

\subsubsection{Skill Embedding and Skill Prior Learning}\label{app:generative}

In this section, we provide a detailed overview of the network architectures used to learn the two different skill modules in this work. The training dataset consists of sequences of state-action pairs (skills) collected by running the robot in the environment. We detail the data collection process in Appendix \ref{app:data}. For the Fetch push, cleanup and stacking tasks, the state information consisted of a 19-dimensional vector that describes the configuration of the robot and the pose of the red block in the environment. For the hook task, we additionally concatenate the pose of the hook and its velocity to give a resulting state vector of 43-dimensions. The action space for all tasks consists of a 4-dimensional vector that encapsulates the agent's end-effector pose in 3-dimensional space and the gripper position. The skill-horizon $H$ was set as 10 across all environments.

We parameterise both the VAE skill embedding and normalising flows skill prior modules as deep neural networks and detail their implementation below.

\paragraph{VAE embedding module} The VAE encoder first processes individual concatenated state-action pairs in sequence using a one-layer LSTM with 128 hidden units. The hidden layer after processing the final observation is then fed into an MLP block comprised of 3 linear layers with batch normalisation and ReLU activation units with two output heads over the last layer, yielding parameters $\mu_z$, $\sigma_z$ of the variational posterior $\log q_\phi(z\mid\textit{\textbf{s}}, \textit{\textbf{a}}) \sim \mathcal{N}(\mu_z, \sigma_z)$. We set our latent space to be 4-dimensional, i.e.~$\mathcal{Z}\cong \mathbb{R}^4$. The decoder network mirrors the architecture of the MLP block used in the encoder, taking as input concatenated latent skill vector $z$ and current state $s_t$. The final layer has a single head and a \texttt{tanh} layer to ensure actions are bounded between $(-1, 1)$. The full action sequence ${\textit{\textbf{a'}}}$ is decoded sequentially, with observed states $s_t$ given to the encoder at each step. The overall loss function for the embedding module is as provided in \eqref{eq:vaerecon}, noting the expectation is computed by drawing a single sample from posterior $q_\phi(z \mid \textit{\textbf{s}}, \textit{\textbf{a}})$. Furthermore, the VAE reconstruction loss term is the mean-squared error, i.e.~$\log p_\theta\left(a_{t} \mid z, s_{t}\right)\propto (a_t - a'_t)^2 $.

\paragraph{State-conditioned skill prior sampling module} The network parameterising the skill prior $f:\mathcal{Z}\times \mathcal{S} \rightarrow G$ is a conditional real NVP \cite{dinh2016density} which consists of four affine coupling layers, where each coupling layer takes as input the output of the previous coupling layer, and the robot state vector $s_0$ from the start of the skill sequence. We use a standard Gaussian $p_{\mathcal{G}} (g)\sim\mathcal{N}(0,I)$ as our base distribution for our generative model. Our architecture for $f$ is identical to the conditional Real NVP network used by \citet{singh2021parrot}, see \citet{singh2021parrot} for an in-depth explanation. The loss function for a single example is given by
\begin{equation}
    \mathcal{L}_{prior} = \log p_{\mathcal{G}}(f(z, s_0)) + \log \left| \det \frac{\partial f}{\partial z^\top} \right|.
\end{equation}
During training, we pass the initial state vector of the robot before applying the skill as conditioning information. We optimise our model using the Adam optimiser with a learning rate of 1e-4 and batch size of 128. The overall objective for training the skills model for a single training observation is given as
\begin{equation}
    \mathcal{L}_{skills} = \mathcal{L}_{embed} + \mathcal{L}_{prior},
\end{equation}
noting that gradients of the skills prior loss w.r.t.~$z$, i.e.~$\partial \mathcal{L}_{prior}/\partial z$ are blocked. What this means is that the VAE embedding module is trained without being influenced by the skills prior objective, however, the skills prior training is affected because the topology of latent space determined by the embedding module is evolving during training. We found this joint training yielded more expressive skill priors compared to first training on $\mathcal{L}_{embed}$ and subsequently training the skills prior on $\mathcal{L}_{prior}$ after the embedding module has finished training.
% \begin{multline}
%     \mathcal{L}=\sum_{i=1}^{H} \underbrace{\left(a_{i}-\hat{a}_{i}\right)^{2}}_{\text {reconstruction }}-\beta \underbrace{D_{\mathrm{KL}}\left(\mathcal{N}\left(\mu_{z}, \sigma_{z}\right) \| \mathcal{N}(0, I)\right)}_{\text {regularisation}} + \\\underbrace{\sum_{i=1}^{c} \log \left(\left|\operatorname{det}\left(\frac{d n_{i}}{d n_{i-1}}\right)\right|\right)+\log \left(p_{n}(f(\lfloor z \rfloor))\right)}_{\text{prior training}}
% \end{multline}

\subsubsection{Reinforcement Learning Setup}
\label{app:rl}

We jointly train the high-level and low-level policy in a hierarchical manner, where the high-level policy leverages the resulting cumulative reward after a complete skill execution, while the residual component is updated using the reward information generated after every environment step. By jointly training these two systems, we make effective use of all the online experience collected by the agent to update both the high-level and low-level policies, allowing for better overall sample efficiency of downstream learning.

We utilise Proximal Policy Optimisation \cite{schulman2017proximal} as the underlying RL algorithm for both the high and low-level policies, using the standard hyper-parameters given in the SpinningUp implementation \cite{SpinningUp2018} with a clip ratio of 0.2, policy learning rate of 0.0003 and discount factor $\gamma$ = 0.99. We found it important to train the high-level policy without the residual for an initial 20k steps before allowing the residual policy to modify the underlying skills. This allowed the high-level policy to experience the useful skills suggested by the skill-prior without being distorted by the random initial outputs of the residual policy. As opposed to a hard introduction of the residual action, we gradually introduce it by weighting this additive component using a smooth gating function which increases from 0 to 1 over the course of the first 20k steps. We found that this stabilised the training of the hierarchical agent. While there are various ways to schedule this weighting parameter $w$, we utilised the logistic function - with centre $C$ at 10k steps and a growth rate $k$ of 0.0003.

\begin{equation}
    w = \frac{1}{1 + e^{-k(x-C)}}
\end{equation}

\section{Tasks}
\label{app:tasks}

We describe the downstream RL evaluation tasks in detail below. Note that each of the task environments exhibit dynamical and physical variations from the data collection environment used for skill acquisition. 

\noindent\textbf{Slippery Push}\hspace{0.2cm} The agent is required to push a block to a given goal, however, we lowered the friction of the table surface from that seen in the data collection push task. This makes fine-grained control of the block more difficult. The agent receives a reward of 1 only once the block is at the goal location, otherwise, it receives a reward of 0. The task is episodic and terminates after 100 timesteps.

\textbf{Table Cleanup}\hspace{0.2cm} We introduce a rigid tray object in the scene, into which the agent must place the given block. The tray was not present in the data collection environment and the edges act as an obstacle that the downstream agent must overcome with its available skills. The agent receives a reward of 1 only once the block is placed in the tray, otherwise, it receives a reward of 0. The task is episodic and terminates after 50 timesteps.

\textbf{Pyramid Stack}\hspace{0.2cm} Stacking task where the agent is required to place a small red block on top of a larger blue block. The prior controllers used for data collection are unable to move the gripper above the height of the larger block. Additionally, this task requires precise placement of the red block. The agent receives a reward of 1 only once the red block is successfully balanced on top of the blue block, otherwise, it receives a reward of 0. The task is episodic and terminates after 50 timesteps.

\textbf{Complex Hook}\hspace{0.2cm} The agent is required to move an object to a target location, however, the robot cannot directly reach the object with its gripper. We introduce a hook that the agent must use to manipulate the object. To add to the complexity of this task, the objects are drawn randomly from an unseen dataset of random objects and the table surface is scattered with random "bumps" that act as rigid obstacles when manipulating the object. The agent receives a reward of 1 only once the block is at the goal location, otherwise, it receives a reward of 0. The task is episodic and terminates after 100 timesteps.

\section{Skill Prior Ablation}
\label{skillpablation}
\begin{figure}[t]
  \centering
  \includegraphics[width=\textwidth]{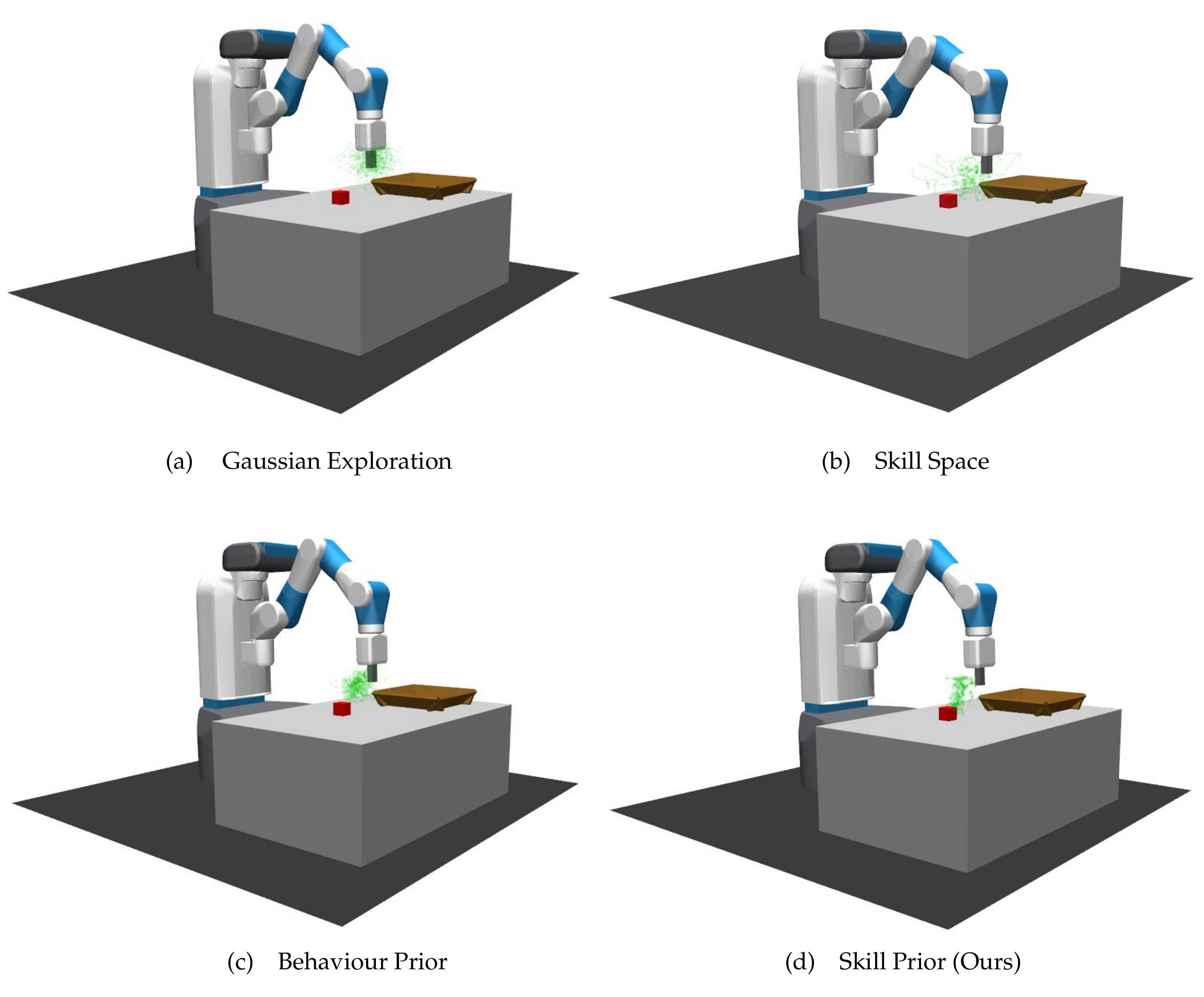}
  \caption{\textbf{Exploratory Trajectories} We plot the trajectories taken by four different strategies used in skill-based and single-step RL approaches. Note how the skill prior significantly directs the exploratory trajectories towards the object in the environment while still allowing the agent to explore a diverse set of surrounding skills.}
  \label{exploration}
\end{figure}

In this section, we analyse the impact that our proposed skill prior has on the exploratory behaviours of the agent during the early stages of training. For a quantitative evaluation, we denote the percentage of the first 20k steps that result in some form of manipulation of objects placed in the scene. The intuition here is that for manipulation-based tasks, "meaningful" behaviours would involve manipulation of the objects in the scene in order for the agent to make progress towards solving the task at hand, as opposed to the extensive random exploration of irrelevant, non-manipulation-based behaviours. We evaluate our skill prior based exploration strategy to approaches used in existing RL literature for both skill-based and standard single-step agents. Table \ref{skillprior_eval} summarises the results. We additionally provide a visual depiction of the trajectories taken by each exploration strategy in Figure \ref{exploration} to better understand how the skill prior impacts exploration.

\begin{table}[t]
\centering
\caption{Skill prior impact on exploration. The table denotes the proportion of exploratory steps that result in some form of manipulation of relevant objects in the environment during the first 20k steps.}
\resizebox{0.9\textwidth}{!}{%
\begin{tabular}{@{}lllll@{}}
\toprule
                   & \textbf{Skill Prior (Ours)}       & Skill Space               &      Behaviour Prior          & Gaussian Exploration \\ \midrule
Object Interaction & \multicolumn{1}{c}{\textbf{45.4\%}} & \multicolumn{1}{c}{9.39\%} & \multicolumn{1}{c}{4.72\%} & \multicolumn{1}{c}{0.560\%} \\ \bottomrule
\end{tabular}%
}
\label{skillprior_eval}
\end{table}

As indicated by the results, our NVP-based skill prior attains the highest proportion ($>$45\%) of meaningful exploratory behaviours and we can attribute this to the ability of the skill prior to sample relevant skills that the agent can directly execute in the environment. We note here that while this prior does heavily bias the agent's behaviours towards manipulating the block, it does not completely constrain the agent's ability to explore other skills that could be potentially better for the downstream task. The Skill Space variant tested in this ablation encompasses skill-based algorithms that sample behaviours from the skill space using the stochastic policy output $\pi_{HL}(z|s)$. This is reminiscent of the strategy used in SPiRL \cite{pertsch2020spirl}, which additionally regularises this distribution towards a learned prior during training. This strategy achieved only 9.39\% of relevant behaviours which we could use to explain the discrepancy in the asymptotic performance of SPiRL and ReSkill (No Residual). We note here that this value would vary based on the strength of the regularisation towards the prior, and should increase as the policy output gets better regularised towards the prior over the course of training. The Behaviour Prior tested in this work is the single-step prior proposed by Singh \textit{et al.} \cite{singh2021parrot} and while it shares a similar operation to our skill prior we note it attains much lower interaction than our skill-based variant. We could attribute this to the temporal nature of skills, which are less noisy than constantly sampling single-step action allowing for more directed exploration towards the block. Figure \ref{exploration} (c) and (d) provides a visual depiction of this. Gaussian exploration has almost no interaction with the block, which significantly impacts the agent's ability to learn as shown in the training curves for SAC, PPO and HAC in Figure \ref{curves}.

\section{Evaluation on Long Horizon Manipulation Tasks}

\begin{figure}
\centering
\begin{minipage}{.5\textwidth}
  \centering
  \includegraphics[width=\linewidth]{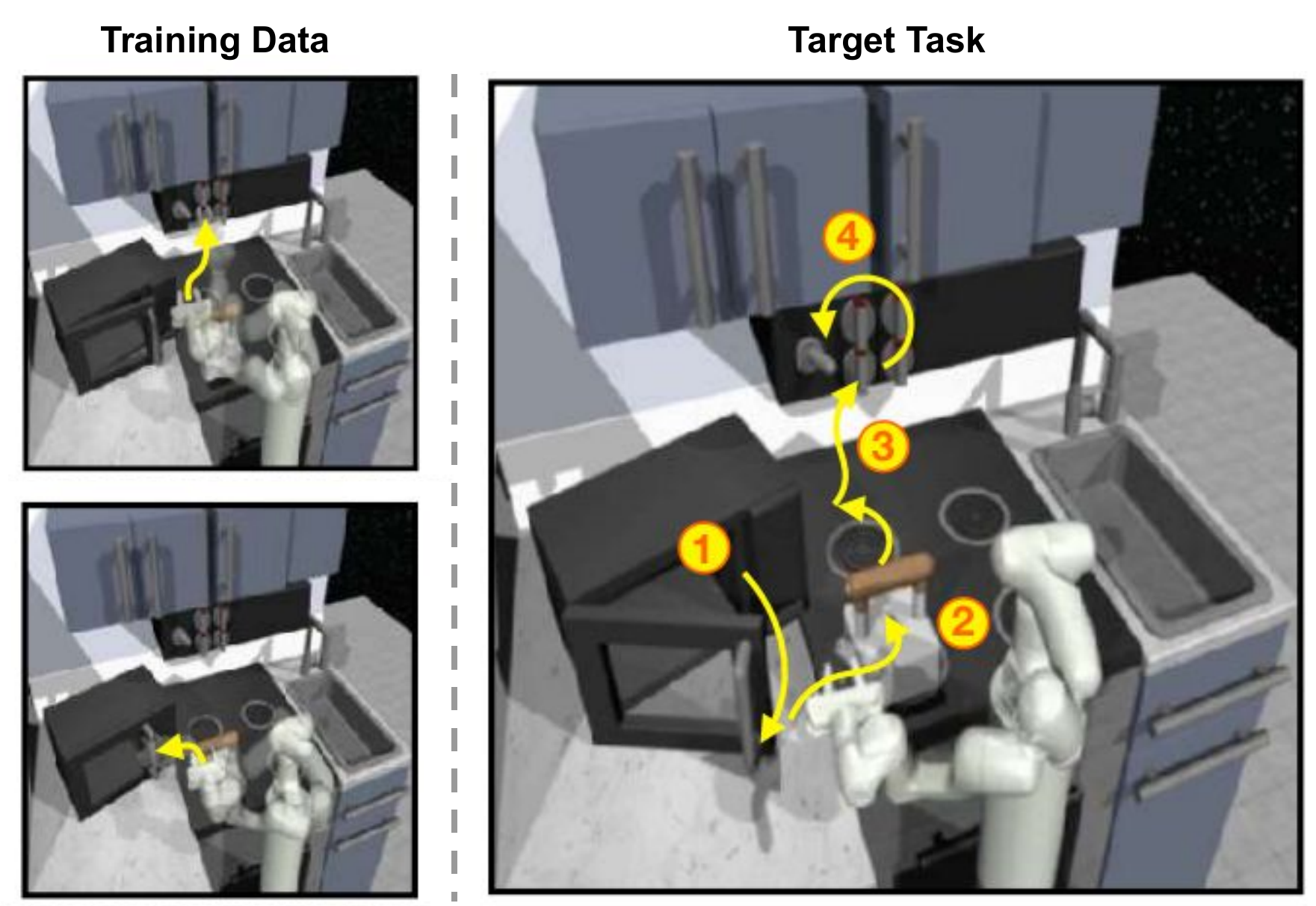}
  \label{fig:test1}
\end{minipage}%
\begin{minipage}{.5\textwidth}
  \centering
  \includegraphics[width=\linewidth]{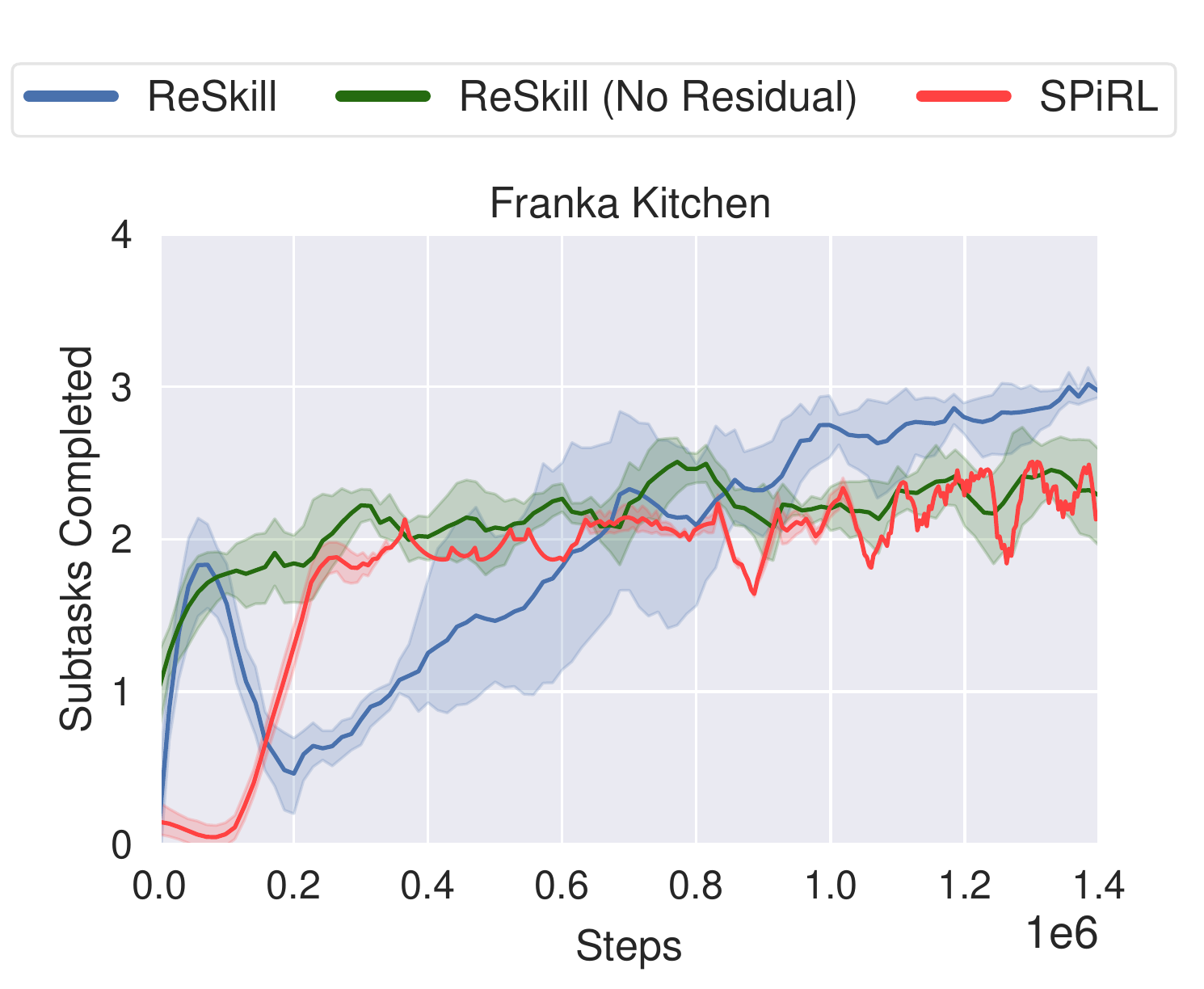}
  \label{fig:test2}
\end{minipage}
\caption{\textbf{Evaluation on Franka Kitchen Environment.} (\textbf{Left}) Franka Kitchen environment proposed by Gupta \textit{et al.} \cite{lynch2019play} which requires the agent to manipulate a kitchen setup to reach a target configuration. (\textbf{Right}) Note how ReSkill exhibits faster convergence to a higher reward than SPiRL.}
\label{kitchen}
\end{figure}

To demonstrate the broad applicability of ReSkill to higher dimensional tasks involving long-horizon goals, we provide an additional evaluation in the Franka Kitchen environment introduced by Gupta \textit{et al.} \cite{lynch2019play}. To train the skills modules, we use the demonstration data provided in the D4RL benchmark \cite{fu2020d4rl}, which consists of 400 teleoperated sequences in which a 7-DoF Franka robot arm manipulates objects in the scene (e.g. switch on the stove, open microwave, slide cabinet door). During downstream learning, the agent has to execute an unseen sequence of multiple sub-tasks. The agent receives a sparse, binary reward for each successfully completed sub-task. The action space of the agent consists of a 7-dimensional vector that corresponds to each of the robot joints as well as a 2-dimensional continuous gripper opening/closing action. The state space consists of a 60-dimensional vector that consists of the agent’s joint velocities as well as the poses of the manipulable objects. We summarise the results in Figure \ref{kitchen}.

The training curves illustrate that ReSkill can effectively handle higher dimensional and long-horizon tasks, and can substantially outperform SPiRL in both sample efficiency and convergence to higher final policy performance. We note the high variance of ReSkill during the gradual introduction of the residual, which we can attribute to the long-horizon nature of the task and the multiple strategies that the low-level agent can identify to adapt to the task at hand. It is important to note however that this agent gradually converges towards a stable solution that can yield a much higher reward after this transition phase. Without the residual, the ReSkill agent attains the same final performance as SPiRL however still demonstrates faster convergence given the direct exploratory guidance provided by our proposed skill prior.

\vspace{0.5cm}
\section{Data Collection}
\label{app:data}

\begin{figure}[h!]
  \centering
  \includegraphics[width=0.8\textwidth]{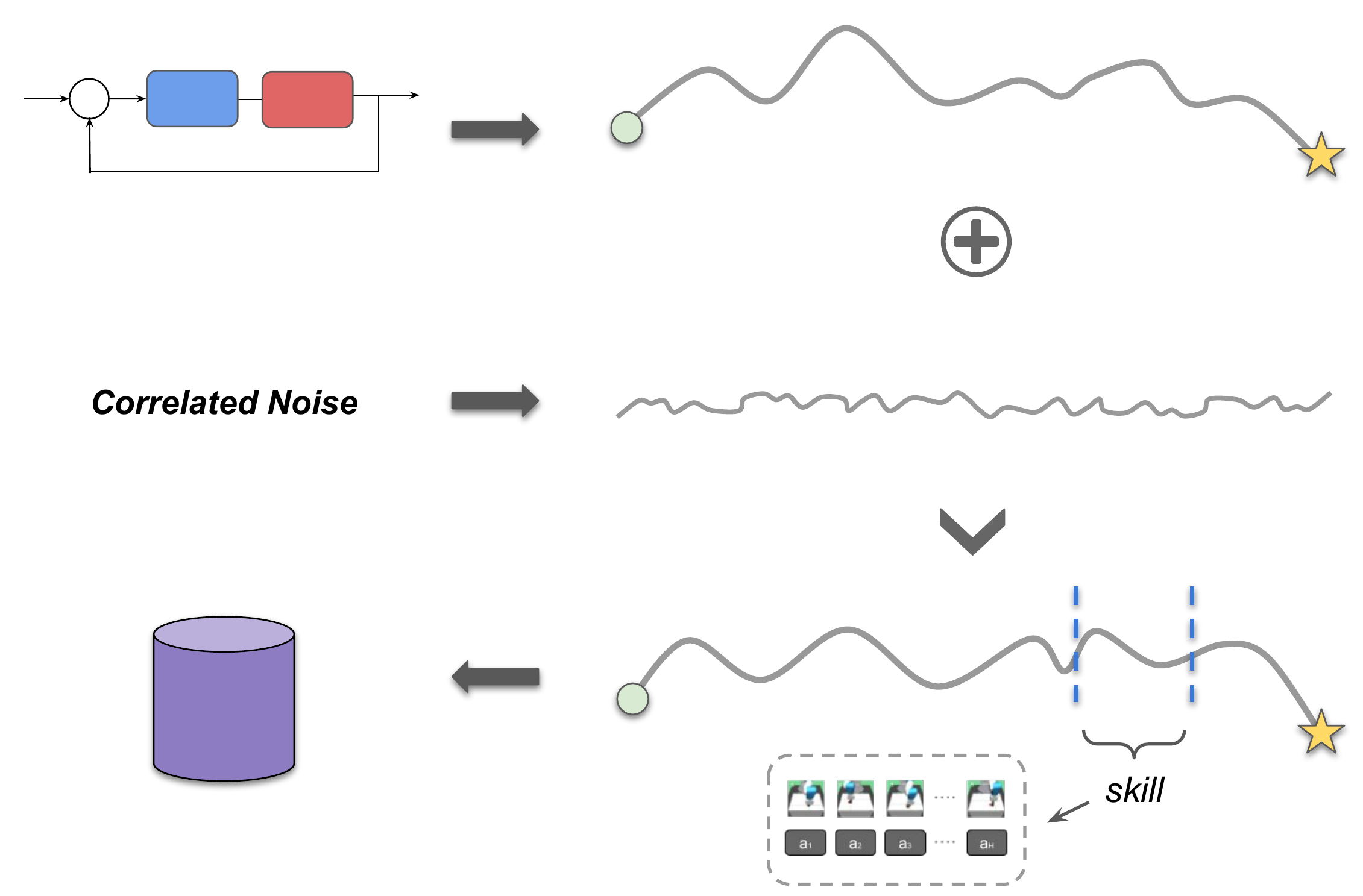}
  \caption{\textbf{Data collection.} Skill extraction from trajectories collected from a handcrafted controller. To obtain a diverse range of skills we add correlated Perlin noise to the trajectory rollouts before skill extraction.}
  \label{skill-extraction}
\end{figure}

In order to re-purpose existing controllers for a wide range of tasks, we decompose their behaviours into task-agnostic skills. We firstly collect a dataset of demonstration trajectories, consisting of state-action pairs by executing the handcrafted controllers in the data collection environment. For the block manipulation tasks, we script 2 simple controllers for a 7-DoF robotic arm each capable of completing \textit{pushing} and \textit{grasping} manipulation tasks on an empty table as described in Algorithm \ref{algorithm2} and \ref{algorithm3}. This dataset was used to train a single skills module that was later used for downstream RL learning across \textit{Slippery-Push}, \textit{Table-Cleanup} and the \textit{Pyramid-Stacking} task. For the hook task, we scripted a simple controller that can manipulate a block using a hook object on an empty table which we used to collect the required dataset. We note here that these controllers are suboptimal with respect to the downstream tasks which each introduce additional complexities to the environment that the RL agent will have to adapt to as described in Appendix \ref{app:tasks}. To increase the diversity of skills collected, we add Perlin noise \cite{perlin1985image} to the controller outputs. Perlin noise is correlated across the trajectory, allowing for smooth deviations in trajectory space. Before adding these trajectories to the dataset, we filter them based on a predefined rule: if it is longer than the skill horizon $H$ we add it to our dataset. Skills can be extracted from a trajectory, by randomly slicing a $H$ dimensional skill consisting of a sequence of actions $\textit{\textbf{a}} = \{a_t,...,a_{t+H-1}\}$ and the corresponding states $\textit{\textbf{s}} = \{s_t,...,s_{t+H-1}\}$ that these actions were executed in. For all experiments conducted in this work, we set the skill horizon $H$ to 10. Figure \ref{skill-extraction} shows a summary of the data collection process and the notion of a skill from a recorded trajectory. We collect a total of 40k trajectories to train the skills module for the Fetch manipulation tasks.

\clearpage

\subsection{Handcrafted Controllers}
\label{app:contr}

We describe each of the controllers derived to complete simple tasks on an empty table with a single object placed in front of it as shown in Figure \ref{robot_envs} (a). This was the environment used for collecting data for training the skill modules.

\subsubsection{Reactive Push Controller}
This controller is designed for pushing an object to a target location. Although this policy performs well in the original Push task, its performance drops dramatically when the sliding friction on the block is reduced as in the \textit{SlipperyPush} task. Pseudo-code for the controller is provided in Algorithm \ref{algorithm2}.

\begin{algorithm}[h!]
    \SetAlgoLined
    \nonl\textbf{Given:} threshold
    
    \KwIn{state}
    \KwOut{action}
    \begin{algorithmic}[1]
    \STATE\textbf{if} distance(targetObjectPose, objectPose) $<$ threshold \textbf{then}\\
        \hspace{0.2cm}action $\leftarrow$ 0
    
    \STATE\textbf{else if} gripper\_at\_pushloc(gripperPose, pushLoc) \textbf{then}\\
        \hspace{0.2cm}action $\leftarrow$ push object to target
    
    \STATE\textbf{else}\\
        \hspace{0.2cm}action $\leftarrow$ move gripper to push location
        
    \STATE\textbf{end if}
        
    \STATE\textbf{return} action
    \caption{Reactive Push}
    \label{algorithm2}
    \end{algorithmic}
\end{algorithm}

\subsubsection{Pick and Place Controller}

This controller is designed to move to an object location, grasp the object and move it towards a target goal. The controller does not lift the block higher than 3cm above the surface of the table and therefore will fail if either the object has to be placed on a higher surface or alternatively, the target is on the other side of a barrier. These failure cases are present in both the \textit{Pyramid-Stack} and \textit{Table-Cleanup} tasks. Pseudo-code for the controller is provided in Algorithm \ref{algorithm3}.

\begin{algorithm}[h!]
\SetAlgoLined
\nonl \textbf{Given:} threshold

\KwIn{state}
\KwOut{action}
\begin{algorithmic}[1]
\STATE\textbf{if} distance(targetObjectPose, objectPose) $<$ threshold \textbf{then}\\
    \hspace{0.2cm} action $\leftarrow$ 0

\STATE\textbf{else if} is\_grasped(gripperPose, objectPose) \textbf{then}\\
    \hspace{0.2cm}action $\leftarrow$ move to target

\STATE\textbf{else if} object\_in\_gripper(gripperPose, objectPose) \textbf{then}\\
    \hspace{0.2cm}action $\leftarrow$ close gripper

\STATE\textbf{else if} gripper\_above\_object(gripperPose, objectPose) \textbf{then}\\
    \hspace{0.2cm}action $\leftarrow$ move gripper down
    
\STATE\textbf{else}\\
    \hspace{0.2cm}action $\leftarrow$ move above object
    
\STATE\textbf{end if}
    
\STATE\textbf{return} action
\end{algorithmic}
\caption{Pick and Place}
\label{algorithm3}
\end{algorithm}

\clearpage
\subsubsection{Reactive Hook Controller}

This controller is designed to pick up a hook, move it behind and to the right of an object and pull the object towards a target location. The controller and environment are adapted from the set of tasks presented by Silver \textit{et al.} \cite{silver2018residual}. The controller works well when the table is empty and when the object being manipulated is a simple block. In the \textit{Complex-Hook} environment, however, this controller falls suboptimal as we introduce additional objects of varying masses and shapes as well as rigid "bumps" which make the surface of the table uneven. This causes the hook to get stuck when sliding objects along the table. Pseudo-code for the controller is provided in Algorithm \ref{algorithm4}.

\begin{algorithm}[h!]
\SetAlgoLined
\setcounter{AlgoLine}{0}
\nonl\textbf{Given:} threshold

\KwIn{state}
\KwOut{action}
\begin{algorithmic}[1]
\STATE\textbf{if} distance(targetObjectPose, objectPose) $<$ threshold \textbf{then}\\
    \hspace{0.2cm}action $\leftarrow$ 0

\STATE\textbf{else if} hook\_is\_not\_grasped(gripperPose, hookPose) \textbf{then}\\
    \hspace{0.2cm}action $\leftarrow$ grasp hook

\STATE\textbf{else if} hook\_in\_position(hookPose, objectPose) \textbf{then}\\
    \hspace{0.2cm}action $\leftarrow$ place hook to the right of object
    
\STATE\textbf{else}\\ 
    \hspace{0.2cm}action $\leftarrow$ move object to target

\STATE\textbf{end if}
    
\STATE\textbf{return} action\end{algorithmic}
\caption{Reactive Hook}
\label{algorithm4}
\end{algorithm}

\end{document}